\documentclass[10pt,twocolumn,letterpaper]{article}

\usepackage{iccv}
\usepackage{times}
\usepackage{epsfig}
\usepackage{graphicx}
\usepackage{amsmath}
\usepackage{amssymb}

\usepackage{caption}
\usepackage{multirow}
\usepackage{makecell}
\usepackage{graphics}
\usepackage{afterpage}
\usepackage{xcite}
\usepackage[normalem]{ulem}

\usepackage{xspace}
\usepackage{bbold}
\usepackage{xcolor}

\usepackage{booktabs}

\usepackage[pagebackref=true,breaklinks=true,letterpaper=true,colorlinks,bookmarks=false]{hyperref}

\iccvfinalcopy 
\def\iccvPaperID{0000} 

\ificcvfinal\fi

\begin{document}

\title{Are we Missing Confidence in Pseudo-LiDAR Methods\\for Monocular 3D Object Detection?}

\author{Andrea Simonelli$^1$, Samuel Rota Bul\`{o}$^2$, Lorenzo Porzi$^2$, Peter Kontschieder$^2$, Elisa Ricci$^1$\\
 $^1$University of Trento, $^1$Fondazione Bruno Kessler, $^2$Facebook Reality Labs
}

\newcommand{\ourmethod}{Pseudo-MonoDIS\xspace}
\newcommand{\pl}{PL\xspace}
\newcommand{\eli}[1]{{\color{magenta}#1}}

\newcommand{\myparagraph}[1]{
\vspace{3pt}\noindent\textbf{#1}
}

\maketitle
\ificcvfinal\thispagestyle{empty}\fi

\begin{abstract}
{Pseudo-LiDAR-based methods for monocular 3D object detection have received considerable attention in the community due to the performance gains exhibited on the KITTI3D benchmark, in particular on the commonly reported validation split.
This generated a distorted impression about the superiority of Pseudo-LiDAR-based (PL-based) approaches over
methods working with RGB images only.
Our first contribution consists in rectifying this view by 
pointing out and showing experimentally that the validation results published by PL-based methods are substantially biased.
The source of the bias resides in an overlap between the KITTI3D object detection validation set and the training/validation sets used to train depth predictors feeding PL-based methods. Surprisingly, the bias remains also after geographically removing the overlap.
This leaves the test set as the only reliable set for comparison, where published PL-based methods do not excel.
Our second contribution brings PL-based methods back up in the ranking with the design of a novel deep architecture which introduces a 3D confidence prediction module. We show that 3D confidence estimation techniques derived from RGB-only 3D detection approaches can be successfully integrated into our framework and, more importantly, that improved performance can be obtained with a newly designed 3D confidence measure, leading to state-of-the-art performance on the KITTI3D benchmark.}
\end{abstract}
\vspace{-10pt}

\section{Introduction}\label{sec:intro}
By providing information about pose, location and category of objects in the 3D space,
3D object detection constitutes an enabling technology for applications like autonomous driving or augmented reality. 
To obtain accurate localisation performance, existing solutions rely on depth information inferred from stereo cameras or derived from Light Detection and Ranging (LiDAR) sensors. The downsides of both variants are an increase of costs, the necessity of involved recalibration routines and the inhibition of the product design form factors due to 
fabrication constraints. 

\begin{figure}[t]
    \centering
    \includegraphics[width=\columnwidth]{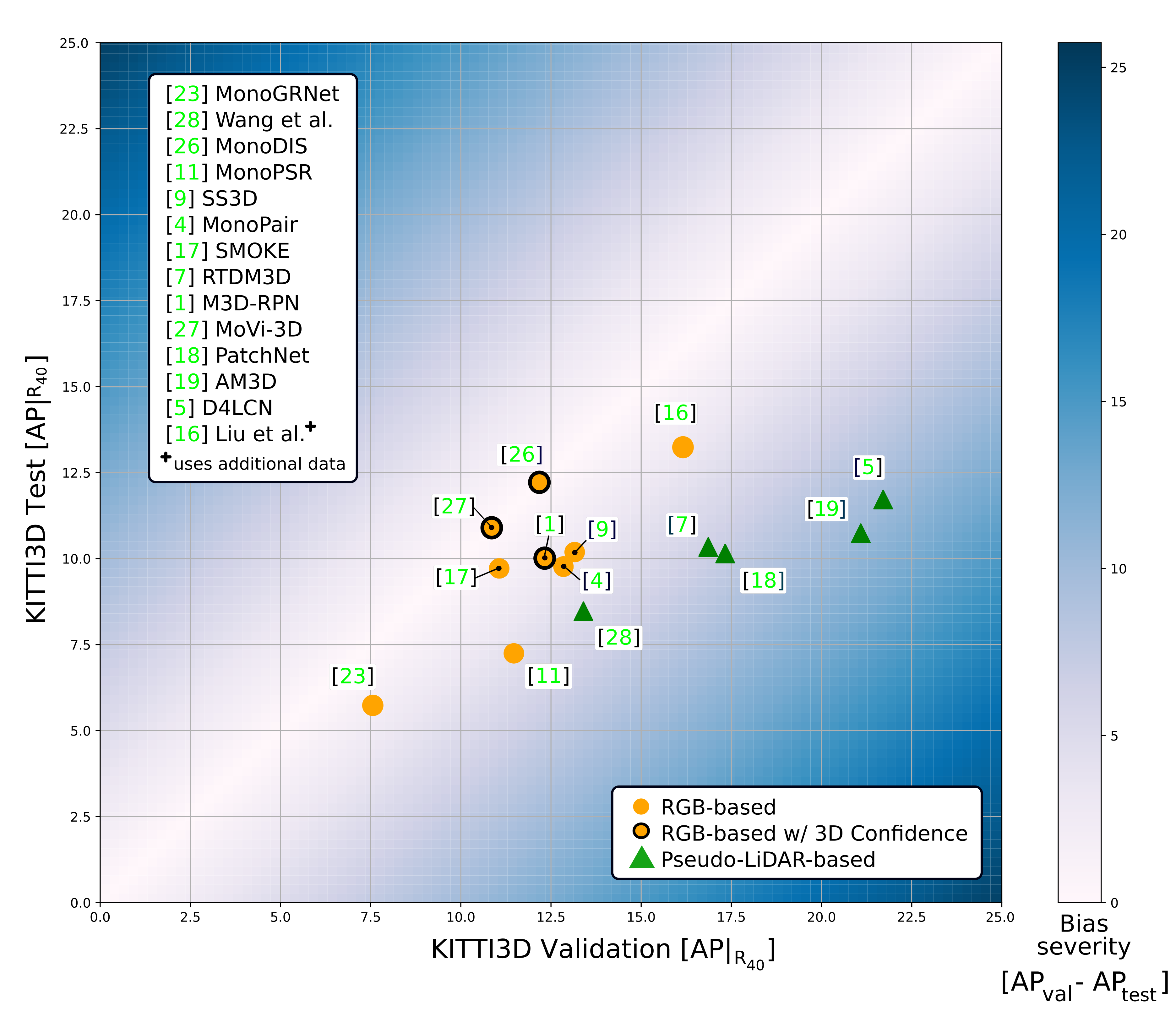}
    \caption{
    Performance of state-of-the-art 3D detection methods on the KITTI3D validation and test sets\textcolor{red}{$^1$}. RGB-based methods (orange circles) exhibit a low performance discrepancy between the two sets, whereas Pseudo-LiDAR-based methods (green triangles)  perform much better (up to 10 AP) on validation than on test. This indicates a bias, which we display by means of a blue-toned colormap.  
    These results also show that the best performing RGB-based methods generally benefit from exploiting a \textit{3D Confidence} (circled orange circles), a component which has not yet been introduced in any PL-based methods.
    }
    \label{fig:teaser}
    \vspace{-10pt}
\end{figure}

\footnotetext[1]{We took as reference the performance on class \textit{Car} in the \textit{Moderate} difficulty, computed with the $AP|_{R_{40}}$ metric \ie the one used as reference on the official KITTI3D benchmark.}

To overcome 
these issues, an emerging branch of 3D object detection methods is entirely based on monocular cameras \cite{brazil2019m3d, jorgensen2019ss3d, ma2019am3d, Manhardt_2019_CVPR, simo2019monodis, simo2020virtual, wang2019pseudo}. Monocular cameras are a cheap alternative to the expensive LiDAR or stereo setups, but, at the same time, incur a substantially increased algorithmic complexity due to the absence of depth observations. Indeed, accurate estimation of an objects' distance to the camera is the most difficult task in monocular, image-based 3D object detection, 
making it an ill-posed problem. 
Despite the development of methods which focus on increasing the generalization with respect to distance \cite{brazil2019m3d, simo2020virtual}, monocular image-based methods still lag far behind their LiDAR or stereo-based counterparts. 

A recent line of works~\cite{Manhardt_2019_CVPR,Xu_2018_CVPR} has leveraged Convolutional Neural Networks (CNNs) for image-based depth predictions as depth substitute in monocular 3D object detection algorithms. Pseudo-LiDAR (\pl)~\cite{wang2019pseudo, yang2019pseudopp} was promoted as a particularly effective depth representation, reporting impressive results on the challenging KITTI3D benchmark \cite{Geiger2012CVPR}. It essentially mimics a LiDAR signal for a RGB image by projecting each 2D pixel from its corresponding, estimated depth map into 3D space. With the resulting 3D point cloud, the 3D detection task is usually approached by applying {state-of-the-art} LiDAR-based (and thus 3D point-based) detection algorithms. PatchNet~\cite{Ma_2020_rethink} has recently refuted 3D points as the source of \pl's effectiveness by providing an equivalently performing implementation based on stacking 3D world coordinates as 2D maps. While this eliminates the claims of \pl being advantageous due to its 3D point-based representation, their ablations confirmed the importance of operating on transformed 2D image coordinates incorporating camera intrinsics (focal length and principal point). 

In this paper we argue that \pl-based approaches, and more in general approaches that take depth as input, have introduced a distorted perception in the research community about their performance in the monocular setting with respect to other state-of-the-art methods that use RGB-images only. We identified two main reasons behind the issue, which constitute the two main contributions of this paper. 

\myparagraph{First contribution.}
State of the art \pl-based methods report excellent performance on the KITTI3D validation set but do not show the same gains on the test set. In this work we perform an in-depth experimental study to analyze the reasons behind such inconsistency and demonstrate that top performing \pl-based methods adopt a training protocol which artificially leads to high average precision on the validation set. The issue is evident in Fig.~\ref{fig:teaser}, where the discrepancy between the KITTI3D validation and test set performance of \pl-based methods (green triangles) is much more pronounced than RGB-based methods (orange circles).
Indeed, the depth estimation algorithms on which \pl-based methods heavily rely are usually trained by including $\approx30\%$ of the validation set data used for 3D object detection. Despite this issue was mentioned briefly in~\cite{wang2019pseudo}, this biased training protocol was later used in many subsequent \pl-based methods. This clearly indicates the necessity and the relevance for the community of a more detailed analysis, which we provide in this paper.

\myparagraph{Second contribution.} The outcome of a fair comparison on test set of \pl-based methods against RGB-only based approaches 
on the KITTI3D benchmark is currently favouring more the latter ones. On the flip side, we found that published \pl-based methods are penalized by the complete lack of a proper 3D confidence score which, as shown in Fig.~\ref{fig:teaser} (circled orange circles), is becoming a fundamental component of state-of-the-art RGB-only methods.
In this paper we propose, for the first time, to endow \pl-based methods with a mechanism for predicting a 3D confidence, demonstrating remarkable performance gains. 
In particular, we show that, following previous RGB-only based methods \cite{simo2019monodis}, also in the case of \pl, 3D confidence can be trained by directly regressing the expected loss. While this works well in practice, it is sensitive to the scale of the loss and, hence, requires some hyperparameter tuning. Moreover, it suffers from the issue of becoming overconfident as the training progresses towards an overfitting regime. In the spirit of addressing those two issues, we open a novel direction and successfully explore the possibility of having 3D confidences expressed in relative terms. Our novel finding leads to improved performance and sets the new state-of-the-art on the KITTI3D benchmark.

\section{Related Works}

Current approaches for monocular 3D object detection can be roughly divided in two categories: RGB-only methods, which directly address the ill-posed problem of the object's distance estimation, and \pl-based methods, which leverage from automatically estimated depth maps or point clouds to recover the distance information.

\myparagraph{Monocular RGB-only 3D detectors.} Earlier approaches for monocular RGB-only 3D detection such as SSD-6D~\cite{Kehl_2017_ICCV} and Deep3DBox~\cite{Mousavian_2017_CVPR} build on top of state of the art deep architectures for 2D detection, and exploit information from projective geometry to estimate the 3D pose and position of the objects in the scene.
Mono3D~\cite{Chen_2016_CVPR} develops from the idea of generating 3D proposals and scoring them according to several cues, such as semantic segmentation features, object contour, and location priors.
OFTNet~\cite{Roddick18} operates by considering an orthographic feature transform to map a 2D feature map to bird-eye view. 
MonoGRNet~\cite{qin2019monogrnet} 
simultaneously estimates 2D bounding boxes, instance depth, 3D location of objects and local corners.
GS3D~\cite{li2019gs3d} exploits an off-the-shelf 2D object detector and efficiently computes a coarse cuboid for each predicted 2D box, which is then refined to estimate the 3D bounding box. MonoPSR~\cite{ku2019monopsr} jointly leverages 3D proposals and scale and shape estimation to accurately predict 3D bounding boxes from 2D ones. Recently, few works have proposed single-stage deep architectures \cite{brazil2019m3d, simo2020virtual}. M3D-RPN \cite{brazil2019m3d} generates 2D and 3D object proposals simultaneously and exploits a post-processing optimisation and a depth-aware network to improve localization accuracy. MoVi-3D \cite{simo2020virtual} is a lightweight architecture which exploits automatically generated virtual views where the object appearance is normalized with respect to distance to facilitate the detection task.
Liu \etal~\cite{liu2020smoke} propose SMOKE, a deep architectures which predicts 3D bounding boxes by relying on key-point estimation as an intermediate task. MonoDIS~\cite{simo2019monodis} shows that training convergence and detection accuracy of 3D detection networks can be improved by considering loss disentanglement. In~\cite{simo2019monodis} 3D confidence for detection is also introduced for increasing performance. In this paper we show how this notion can be extended to \pl and further improved introducing a relative measure of confidence.

\myparagraph{Pseudo-LiDAR based 3D detectors.}
A second category of works exploit external data and network models to generate depth maps from the RGB input as an intermediate step for 3D detection.
For instance, ROI-10D~\cite{Manhardt_2019_CVPR} introduces a loss to minimize the misalignment of 3D bounding boxes and exploits depth maps inferred with SuperDepth~\cite{Pillai_2019_ICRA}. A disparity prediction module is considered in~\cite{Xu_2018_CVPR} and integrated into a network composed of two parts: one that generates 2D region proposals, and another that predicts 3D object location, size and orientation.
Pseudo-Lidar~\cite{wang2019pseudo} represents the first \pl method, introducing the idea of interpreting depth maps as 3D point clouds which are then fed to state-of-the-art LiDAR-based 3D object detectors. In \cite{wang2019pseudo} the presence of a possible performance bias is also suggested but an in-depth experimental study is lacking. 
Pseudo-Lidar++~\cite{yang2019pseudopp} improves the accuracy in the localisation of faraway objects by adapting a stereo network architecture and deriving a loss function for direct depth estimation. 
AM3D \cite{ma2019am3d} proposes to integrate complementary RGB features into the \pl pipeline and introduces a a specific module to map the 2D image data
to the 3D point cloud.
PatchNet \cite{Ma_2020_rethink} analyses the effect of depth data representation on performances and improves over previous \pl models by integrating the 3D coordinates as additional channels of input data.
However, all these works lack a fundamental component of state-of-the-art RGB-only based detectors \cite{simo2019monodis} \ie the estimation of a 3D confidence.

\section{Preliminaries}
We first review the monocular 3D object detection task and introduce the KITTI dataset~\cite{Geiger2012CVPR} -- the most influential benchmark to assess the performance of 3D detection methods. We also report the results of an experimental analysis on KITTI, highlighting the crucial role of depth estimation on the performance of state-of-the-art \pl-based methods.

\subsection{Monocular 3D Object Detection}\label{sec:3d_detection}
The monocular 3D object detection task consists in detecting and localizing all the visible objects of interest (\eg cars) by means of 3D bounding boxes given a single RGB image as input.
Localization must be done in 3D space, properly estimating the 3D coordinates (in meters) of the center of the object $O_i=(X_i, Y_i, Z_i)$, where $X_i, Y_i$ are related to the horizontal and vertical translations, respectively, and $Z_i$ is the distance of the object's center from the camera. The localization also includes the estimation of the object's metric shape $S_i=(H_i, W_i, L_i)$ representing the object's height, width and length, as well as the object's rotation $R_i$ \wrt the camera reference system. The detection requires also to estimate a confidence value $C_i$ which generally reflects the quality and determines how confident the detector is about the particular 3D detection. In this monocular setting, it is common to assume to have a calibrated camera and to know the corresponding intrinsic camera parameters.

\subsection{The KITTI Dataset}\label{sec:kitti}
The KITTI Dataset comprises a broad collection of data from street-level sequences, captured with a multi-sensor setup in the city of Karlsruhe (Germany) in 2011. The remarkable diversity of the sensors enabled many benchmarks, including \textit{3D object detection} and \textit{depth estimation}, which are most relevant for this work. 

\myparagraph{KITTI 3D object detection benchmark (KITTI3D).}
To our knowledge, all 3D object detection methods, and in particular monocular image-based ones, adopted KITTI3D as their predominant, and usually exclusive, testing field. The KITTI3D benchmark is composed of an official \textit{training} and \textit{testing} split, comprising 7481 and 7518 images, respectively. Following \textit{Chen \etal}~\cite{chen2015validationdetection}, it is common to split the training set into \textit{unofficial} training and validation splits, with 3712 and 3769 images, respectively. KITTI provides 2D and 3D bounding box annotations for \textit{Cars}, \textit{Pedestrians} and \textit{Cyclists}, and each box is assigned to one of the \textit{difficulty} levels \textit{Easy}, \textit{Moderate} or \textit{Hard}, depending on the object's 2D height ($\approx$~object's distance), degree of occlusion, and truncation.
KITTI3D adopts two main evaluation metrics, \ie, \textit{3D Average Precision (3D AP)} and \textit{Bird's Eye View Average Precision (BEV AP)}. As reported in~\cite{simo2020pami}, AP$|_{R_{40}}$ is the only legitimate 3D detection AP score, deprecating the previously used AP$|_{R_{11}}$ score.

\myparagraph{Depth prediction benchmark.}
The KITTI depth prediction benchmark offers \textit{official} training and testing splits, but it is common to split~\cite{Eigen2015} the training data into \textit{unofficial} training and validation sets of 23488 and 697 images, respectively. Depth prediction methods are inferring pixel-specific distance estimates \wrt the camera and are evaluated with several metrics like \textit{Absolute Relative Error (AbsRel)}, \textit{Squared Relative Error (SqRel)}, \etc.

\subsection{The Crucial Role of Depth}\label{sec:oracle}
We also provide the results of an \textit{oracle} analysis demonstrating that depth is the most influential factor for performance in monocular 3D object detection.
Following the definitions in Sec.~\ref{sec:3d_detection}, we used KITTI3D predictions of state-of-the-art monocular 3D object detection methods~\cite{brazil2019m3d, Ma_2020_rethink, simo2019monodis,wang2019pseudo} and compared their 3D object detection performances by \textit{substituting} sub-task predictions (\eg \textit{depth}) with their corresponding ground-truth values. 
In Tab.~\ref{tab:biased_oracle} we show that certain sub-tasks like \textit{rotation} ($R$) and \textit{shape} ($W,H,L$) prediction, despite the substitution with ground-truth values, do not significantly improve performance. In contrast, substituting the predicted \textit{depth} estimation ($Z$) with ground truth improves substantially, meaning that \textit{depth} is by-far the most crucial component for 3D object detection. Notably, this observation is consistent for all the different tested methods.

\begin{table}[t]
    \centering
    {\footnotesize
    \resizebox{0.99\columnwidth}{!}{%
    \begin{tabular}{l|c|ccc|ccc}
        \toprule
        
         & Oracle & \multicolumn{3}{c|}{M3D-RPN~\cite{brazil2019m3d}} & \multicolumn{3}{c}{MonoDIS~\cite{simo2020pami}} \\ 
        
        Category & sub-task & Easy & Mod. & Hard & Easy & Mod. & Hard\\ 

        \midrule
        
        \multirow{5}{*}{RGB-based} & -- & 12.78 & 10.36 & 8.07 & 16.71 & 12.32 & 10.58 \\
         & $\hat{R}$ & 14.71 & 11.78 & 9.26 & 17.27 & 12.76 & 11.45 \\
         & $\hat{HWL}$ & 13.47 & 10.52 & 8.26 & 16.75 & 12.56 & 11.29 \\
         & $\hat{XY}$ & 22.63 & 17.47 & 13.48 & 29.59 & 22.17 & 19.31 \\
         & $\hat{Z}$ & \textbf{34.53} & \textbf{28.35} & \textbf{22.51} & \textbf{45.99} & \textbf{38.02} & \textbf{33.48}\\
         
        \midrule
        \midrule

         & Oracle & \multicolumn{3}{c|}{\textit{Wang et al.}~\cite{wang2019pseudo}} & \multicolumn{3}{c}{PatchNet~\cite{Ma_2020_rethink}} \\
        
        Category & sub-task & Easy & Mod. & Hard & Easy & Mod. & Hard\\ 

        \midrule

        \multirow{5}{*}{Pseudo-LiDAR-based} & -- & 23.71 & 12.40 & 10.61 & 31.15 & 16.23 & 13.49\\
         & $\hat{R}$ & 24.04 & 13.39 & 11.13 & 31.60 & 17.43 & 14.58\\
         & $\hat{HWL}$ & 25.73 & 14.50 & 11.64 & 34.19 & 19.01 & 15.58 \\
         & $\hat{XY}$ & 33.76 & 20.37 & 17.22 & 44.23 & 25.62 & 21.76\\
         & $\hat{Z}$ & \textbf{53.71} & \textbf{35.15} & \textbf{29.38} & \textbf{59.81} & \textbf{41.93} & \textbf{35.94}\\

        \bottomrule
        
    \end{tabular}}}
    \caption{
    \textbf{Oracle analyses}. We computed the object detection results (\textit{Car} 3D $AP|_{R_{40}}$) of state-of-the-art methods by substituting selected predicted components (\textit{Oracle}) with their corresponding ground-truth value (\eg $\hat Z$). 
    }
    \label{tab:biased_oracle}
\end{table}

\section{The Bias in Pseudo-LiDAR Experiments}
With depth identified as most critical component in monocular 3D detection works, it becomes obvious that \pl-based methods are particularly sensitive to inputs from depth estimators trained in a biased way. 

\begin{table}[t]
    \centering
    \resizebox{1.0\columnwidth}{!}{
    \begin{tabular}{l|c|ccc|ccc}
        \toprule
        3D Object & Depth & \multicolumn{3}{c|}{\textit{Validation set}~~$3D~AP\uparrow$} & \multicolumn{3}{c}{\textit{Test set~~$3D~AP\uparrow$}}  \\ 

         Detector & Estimator & Easy & Mod. & Hard & Easy & Mod. & Hard \\ 
        \midrule
        \midrule
        
        \textit{Wang et al.} ~\cite{wang2019pseudo} & BTS Eigen    & 24.47 & 13.40 & 10.92 & 9.87 & 6.40 & 5.46 \\
        PatchNet~\cite{Ma_2020_rethink} & BTS Eigen                & 31.60 & 18.22 & 15.10 & 14.00 & 8.70 & 7.39  \\

        \midrule

        \textit{Wang et al.} ~\cite{wang2019pseudo} & BTS GeoSep  & 17.20 &  9.35 &  7.57 & 10.76 & 6.86 & 5.93 \\
        PatchNet~\cite{Ma_2020_rethink} & BTS GeoSep              & 20.79 & 10.55 &  8.90 & 10.88 & 7.42 & 6.51 \\

        \bottomrule
    \end{tabular}}
    \caption{Pseudo-LiDAR results on KITTI3D validation and official benchmark, class \textit{Car}, official $\text{AP}|_{R_{40}}$ metric.}
    \label{tab:detection_bias}
    \vspace{-10pt}
\end{table}

\begin{table}[t]
    \centering
    \resizebox{1.0\columnwidth}{!}{
    \begin{tabular}{l|l|c|c|c|c}
        \toprule
        Train set & Validation Set & $d_1\uparrow$ & $AbsRel\downarrow$ & $RMSE\downarrow$ & $SILog\downarrow$ \\ 
        \midrule

        Eigen & Eigen validation     & 0.908 & 0.084 & 4.003 & 16.577 \\
        Eigen & Detection training   & 0.926 & 0.067 & 3.806 & 15.250 \\
        Eigen & Detection validation & 0.920 & 0.072 & 3.838 & 16.063 \\
        
        \midrule

        GeoSep & GeoSep validation    & 0.904 & 0.093 & 3.627 & 14.019 \\
        GeoSep & Detection training   & 0.858 & 0.111 & 4.830 & 15.960 \\
        GeoSep & Detection validation & 0.872 & 0.105 & 4.429 & 15.872 \\
        
        \bottomrule
        
    \end{tabular}}
    \caption{Depth estimation results with BTS on KITTI, computed \wrt ground-truth depth obtained from LiDAR scans.}
    \label{tab:depth_bias}
    \vspace{-10pt}
\end{table}

\subsection{The Source of the Bias}
To our knowledge, all \pl-based methods published so far were exclusively evaluated on the KITTI3D~\cite{Geiger2012CVPR} dataset which, as described in Sec.~\ref{sec:kitti}, shares data among several benchmarks like 3D object detection and depth prediction. With the advent of \pl, it is however paramount to identify potential sources of cross-pollination in task-specific dataset splits. Our investigations showed that previous, \pl-based works
~\cite{wang2019pseudo, yang2019pseudopp, Ma_2020_rethink} were built on top of DORN~\cite{FuCVPR18-DORN}, \ie a state-of-the-art depth estimator, that in turn however included a majority of images from the detection \textit{validation} set during its training. 
Specifically, we found \textbf{1226/3769} (\textbf{32.5\%}) images to be shared between the widely adopted Eigen~\etal training split~\cite{Eigen2015} for depth estimation and the 
commonly used Chen \etal~\cite{chen2015validationdetection} validation split for 3D object detection. When adding also the images belonging to the same capturing sequence, the numbers slightly increase to 1258/3769 (33.4\%). 

We illustrate the full extent of the contamination in Fig.~\ref{fig:map}, plotting GPS positions and hence the overlap of the different splits (Eigen \etal depth training split in black; Chen \etal validation split for 3D object detection in red). 

In Tab.~\ref{tab:detection_bias}, we show the effect of the contamination on the validation and test scores for two state of the art \pl methods ~\cite{wang2019pseudo,Ma_2020_rethink}. The rows corresponding to \emph{Eigen} are based on biased depth as input, which was generated using BTS~\cite{lee2019bts} trained on the Eigen \etal split. We rely on BTS~\cite{lee2019bts} because it represents a novel state-of-the-art depth estimator\footnote{Differently from the depth estimator usually used by \pl methods, \ie DORN~\cite{FuCVPR18-DORN}, for which official training code is not publicly available, BTS provides a complete open-source code (\url{https://github.com/cogaplex-bts/bts})}. The huge performance drops (up to 17.6 AP) between obtained validation and test scores clearly indicate the relevance of the bias issue discussed here. 

\subsection{Can the Bias be Removed?}\label{sec:unbiased_pl}
As stated above, a contamination exists between the depth training and detection validation sets used by \pl-based methods. To further support our hypothesis that this contamination causes a bias in the KITTI3D validation scores, we introduce a geographical separation between the two data sets. We create a novel depth training set, \textit{GeoSep}, enforcing significant spatial separation between the datasets used for depth estimation and detection. 
Exploiting the GPS information included in the available KITTI benchmark data, we create two novel train/val depth splits by withholding all images i) captured closer than 200m from any KITTI3D training or validation detection image, and ii) belonging to any of the KITTI3D detection sequences.
From a total available amount of 47962 images the aforementioned filtering process yields 22954 images, which we divide in 22287 for the training set and 667 images for the validation set. The distance threshold has been chosen to ensure that our novel \textit{GeoSep} splits would have approximately the same number of images of the Eigen \etal~\cite{Eigen2015} splits, which are 23488 for training and 697 for validation, respectively.
Our new \textit{GeoSep} data split is visualized in Fig.~\ref{fig:map} (green markers), showing a clear safety margin between depth training (Eigen \etal~\cite{Eigen2015}, black markers) and object detection validation (red markers) splits.

\begin{figure}[t]
    \centering
    \resizebox{0.99\columnwidth}{!}{
    \includegraphics[width=\columnwidth]{new_map.pdf}}
    \caption{Geographical distribution of the biased training (black), detection validation (red) and geographically separated (green) depth training splits. Square boxes highlight parts where the overlap between the biased depth training and detection validation sets are particularly evident.}
    \label{fig:map}
    \vspace{-12pt}
\end{figure}

To verify whether our novel split solves the bias issue, we use it to train a depth estimation model. We use BTS~\cite{lee2019bts} as depth prediction network and, again, consider the state-of-the-art \pl methods in~\cite{wang2019pseudo} and~\cite{Ma_2020_rethink}. The results of our analysis, shown in Tab.~\ref{tab:depth_bias} and Tab.~\ref{tab:detection_bias}, still indicate the presence of a bias in both depth estimation and 3D detection results. 
To our great surprise, despite the lack of geographical intersection between the training splits, the gap between validation and test results is still substantially higher (up to 10 AP) compared to the gap that methods using RGB-only inputs typically incur ($\approx$ 3-5 AP).
This suggests a more structured form of contamination that goes beyond the simple geographical distribution of the data, perhaps related to intrinsic factors such as the visual appearance and semantic similarity of the scenes (\eg presence of similar streets). 

\begin{figure}[h!]
    \centering
    \includegraphics[width=0.99\columnwidth]{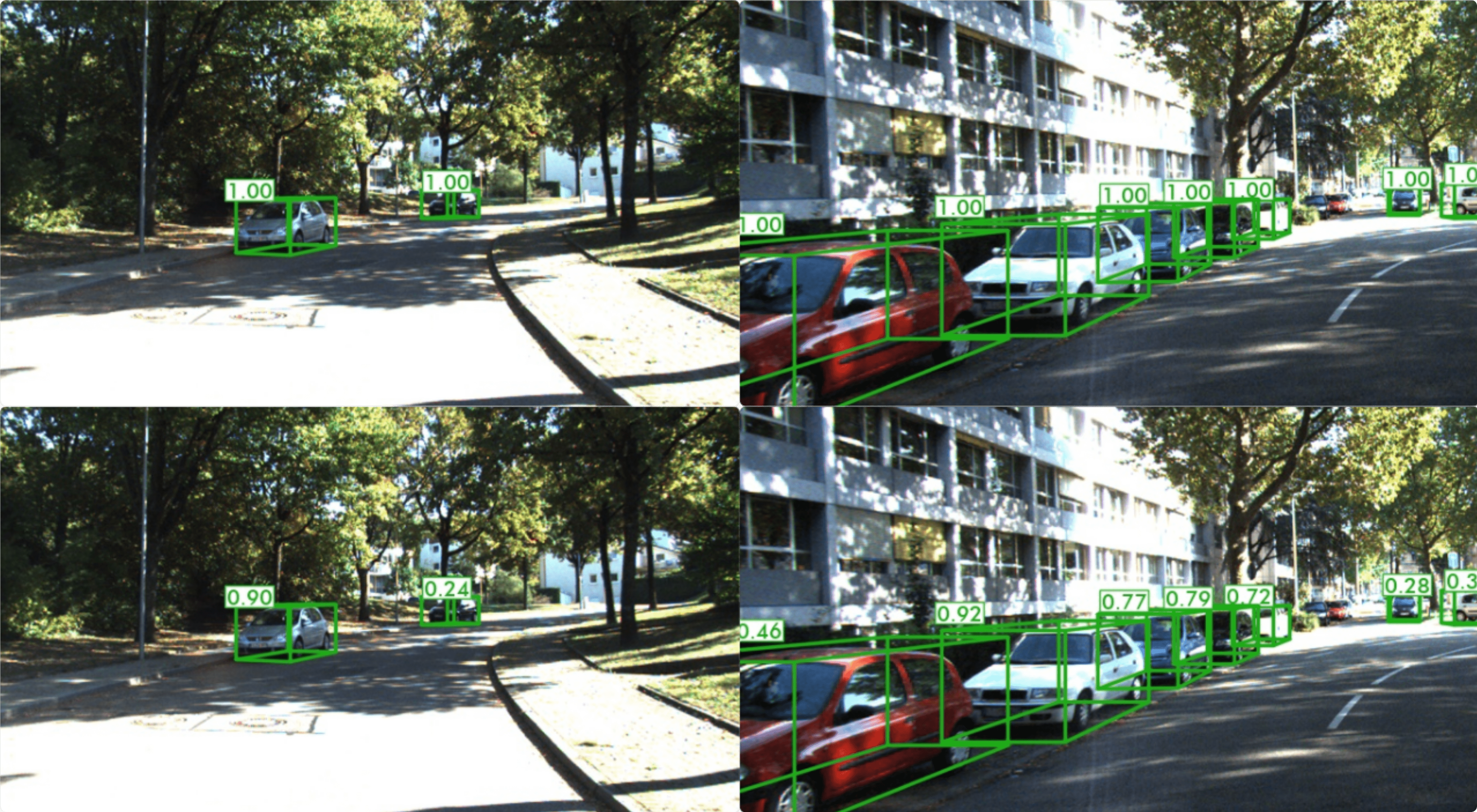}
    \caption{Qualitative results of our method with confidence scores of each detection. Top: We report the 2D confidence score that PL-based methods typically use. Bottom: We report the learned 3D confidence predicted by our method.}
\label{fig:detections}
\end{figure}

The persistence of the bias using both depth training splits (Eigen or \textit{GeoSep}) make us draw the conclusion that fair comparisons should not, at least in these settings, be performed on the KITTI3D validation set.
On the other hand, the fact that published \pl-based methods are not able to surpass state-of-the-art RGB-only based methods (see Tab.~\ref{tab:kitti_test_car} first block) is an indication that the test set itself does not suffer from the same type of bias, thus preserving its validity for the sake of fair comparisons.
Following these conclusions, all the comparisons related to our second contribution will be made on the official test set while the validation set will be used for ablation studies.
Despite our study only partially identifies the source of the bias, this work provides the first analysis of the issue revealing potentially unfair comparisons and we encourage the community to take it into account for future works. 

\begin{figure*}[t]
    \centering
    \includegraphics[width=1.99\columnwidth]{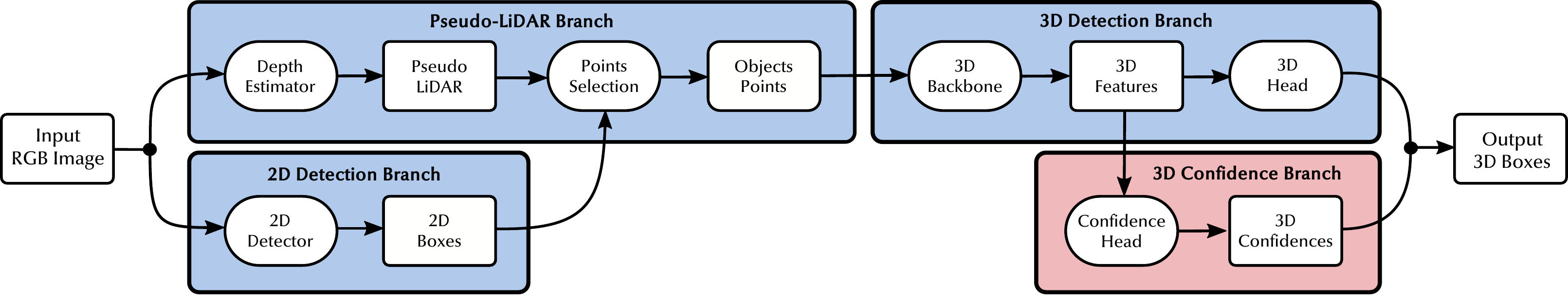}
    \caption{Architecture of a generic Pseudo-LiDAR-based method integrating the proposed 3D confidence component.}
    \label{fig:pipeline_all}
    \vspace{-12pt}
\end{figure*}

\section{3D Confidence for \pl-based Methods}

As described in the previous section, the performance of \pl-based methods is deeply influenced by the upstream depth estimation task. We will now demonstrate that the estimation of the 3D confidence has an equally relevant role. 

The 3D confidence can be thought as an estimate of the quality of the 3D detection which, as described in Sec.~\ref{sec:3d_detection}, has to be associated to each 3D bounding box. In datasets such as KITTI3D, this confidence takes an active role in the computation of the metrics (\eg Average Precision). In light of this fact we observed that existing Pseudo-LiDAR methods 
do not perform the 3D confidence estimation in any way but rely on the class probability coming along with the 2D detections.
By doing so, the confidence adopted by current \pl-based methods is actually agnostic to the quality of the 3D predictions and therefore not effective for the role it should take.
On top of this, as shown in Fig.~\ref{fig:detections}, we observed that 2D detectors are often too confident and therefore the need for a 3D confidence seems essential. 
For this reason, we propose to endow \pl-based methods with the ability of estimating the 3D confidence.

\subsection{Proposed Architecture}
In order to describe how the confidence is estimated we first provide an overview of the general architecture we adopt, similarly to other \pl-based methods \cite{wang2019pseudo, Ma_2020_rethink}, and subsequently detail our contribution.

\myparagraph{PL-based 3D detection architecture.}
The architecture commonly adopted by state-of-the-art \pl-based methods, which we also use in this work, is depicted in Fig.~\ref{fig:pipeline_all} (excluding the red block, \ie our contribution). 
It can be divided into three main branches, namely \textit{2D Detection}, \textit{Pseudo-LiDAR} and \textit{3D Detection}. 
The \textit{2D Detection} and \textit{Pseudo-LiDAR} components usually exploit pre-trained architectures and have the purpose of understanding where the object of interest are in the image, as well as of estimating the per-pixel depth, respectively. The per-pixel depth-map is then converted to Pseudo-LiDAR 3D point-cloud and, finally, the points belonging to each object are selected and filtered to discard elements corresponding to \eg road, occlusions. 
The 3D detection block is responsible for the estimation of the output 3D bounding boxes, taking as input the selected \pl points to perform a point-based 3D detection by means of an initial \textit{3D Backbone} followed by a \textit{3D Head}. 

\myparagraph{3D confidence head.}\label{sec:conf_head} In the following we describe our main contribution, \textit{i.e.} an approach which endows the \pl-based methods under consideration with the ability to predict a self-supervised 3D confidence.
In order to reliably and accurately estimate the 3D confidence of bounding boxes, appropriate 3D-related feature representation need to be computed. For this reason, in this work we introduce an additional branch in the architecture, namely the \textit{3D Confidence Branch}, which, as shown in Fig.~\ref{fig:pipeline_all} (red block), takes as input the set of \textit{3D Features} computed by the \textit{3D Backbone} and outputs a single value $C_i$, \ie the 3D confidence, for each object. In the presence of $K>1$ classes the output is a set of $K$ confidences $C_i^k$, one for each class $k$.
Note that our proposed \textit{3D Confidence Branch} is not tied to any particular architecture and requires minimal modifications to existing \pl approaches. An example of a simple implementation is by mirroring the architecture of the \textit{3D Head}, thus leading to limited computational complexity and minimal overhead in term of inference time.

\subsection{Learning the 3D Confidence}\label{sec:learn_conf}
In this paper we propose two different approaches for 3D confidence prediction. The first approach, which we denote as \textit{Absolute 3D confidence} estimation is inspired by previous RGB-only based methods \cite{simo2020pami}, while the second strategy, called the \textit{Relative 3D confidence} estimation method, is introduced with this paper. In both cases, given a 3D bounding box $B_i$ and the corresponding ground-truth $\hat B_i$, the loss for the 3D confidence prediction $C_i^\text{3D}$ takes the following cross-entropy form:
\[
L_\text{conf}(C^\text{3D}_i|B_i,\hat B_i)=-T_i\log C^\text{3D}_i - (1-T_i)\log(1-C^\text{3D}_i),
\]
where $T_i$ is the target confidence value that takes a different value for the absolute and relative confidences, as described below.
In case of multiple object categories, we assume to have independent 3D confidence predictions per class. 

\myparagraph{Absolute 3D confidence.}
Inspired by \cite{simo2019monodis}, the absolute 3D confidence is trained by directly regressing the loss of the prediction. 
This boils down to setting 
$$T^\text{abs}_i=e^{-\frac{1}{\beta}\ell(B_i,\hat B_i)}$$ as the target confidence, where $\ell(B_i,\hat B_i)$ is the loss incurred by the bounding box prediction and $\beta > 0$ is a temperature parameter. 
Since this approach leads to a 3D confidence which reflects the quality of a 3D detection in absolute terms we call it \textit{Absolute 3D Confidence}. 

\myparagraph{Relative 3D confidence.}
We also propose a novel approach which aims at overcoming a couple of issues that affect loss-based confidences like the one described above. The first is that they are sensitive to the scale of the loss values, which requires to tune scaling factors. The second is that they are not immune to the issue of the network becoming overconfident as the training progresses towards an overfitting regime.
We solve these two issues by shifting the semantics of the score that the network provides with each prediction from an absolute confidence to a relative confidence.
A typical confidence score is supposed to be representative of the absolute quality of the prediction. Instead, the score that we require the network to learn is representative of the quality of the prediction relative to other typical predictions done by the network. For this reason, this novel confidence is regarded as \textit{Relative 3D Confidence}.

\noindent Consider a training set containing $n$ 3D objects, where
$$\ell_i \triangleq \ell(B_i,\hat B_i)$$ denotes the loss incurred by the model on the predicted 3D bounding box for the $i$th object.
We propose to regress as a confidence $C_i^\text{3D}$ for the prediction $B_i$, the proportion of 3D objects in the training set on which the model performs equal or worse than on object $i$, \ie our target confidence prediction is given by
\[
T^\text{rel}_i=\frac{1}{n-1}\sum_{j=1\atop j\neq i}^n {\mathbb 1}_{\ell_j\geq\ell_i}\,,
\]
where $\mathbb 1_P$ denotes the indicator vector for predicate $P$.
The proposed confidence is inherently relative because it does not depend on the actual absolute values of the losses, but rather on their ordering. 
In order to train a model to regress the new confidence, we need to compute the value of $T^\text{rel}_i$ for each 3D object $i$ in the mini-batch. However, to compute such target values, we need to access the loss incurred on each 3D object in the training set, which is computationally demanding. An alternative solution consists in tracking the past loss values for each 3D object in the training set. This is feasible, but given the frequent updates of the model the past losses would be soon outdated. Also, there exist augmentation strategies, for which it might be hard to match predictions across epochs. Interestingly, there is a very simple, stochastic procedure that allows us to train the desired confidence by only accessing loss values in a mini-batch of at least $2$ elements. Specifically, 
we randomly pair the bounding box prediction for 3D object $i$ in the mini-batch with the prediction obtained for another, distinct 3D object $\pi_i$ in the same set. 
Given this assignment, we compute a binary target value $\hat T_i$ for training the confidence as
$ \hat{T_i} = \mathbb{1}_{ \ell_i \leq \ell_{\pi_{i}}},$
and plug this into the cross-entropy loss $L_\text{conf}$ that we use to train the confidence. To see why this works, fix a 3D object $i$ in the training set. Then the target variable $\hat T_i$ is a random variable, where $\pi_i$ is uniformly sampled among the other $n-1$ 3D objects in the training set. Accordingly, the expected value $\mathbb E[\hat T_i]$ of $\hat T_i$ yields exactly $T^\text{rel}_i$. This is also the value to which the predicted confidence $C_i^\text{3D}$ will tend to if trained with the cross-entropy loss $L_\text{conf}$, assuming the losses will eventually converge during training. 
If multiple classes are present, we have independent confidence predictions per class and the random pairing procedure is applied only between 3D objects in the mini-batch having the same class. No loss is computed if a 3D object is the only one of a given class in the mini-batch.

\begin{figure}[t]
    \centering
    \includegraphics[width=0.99\columnwidth]{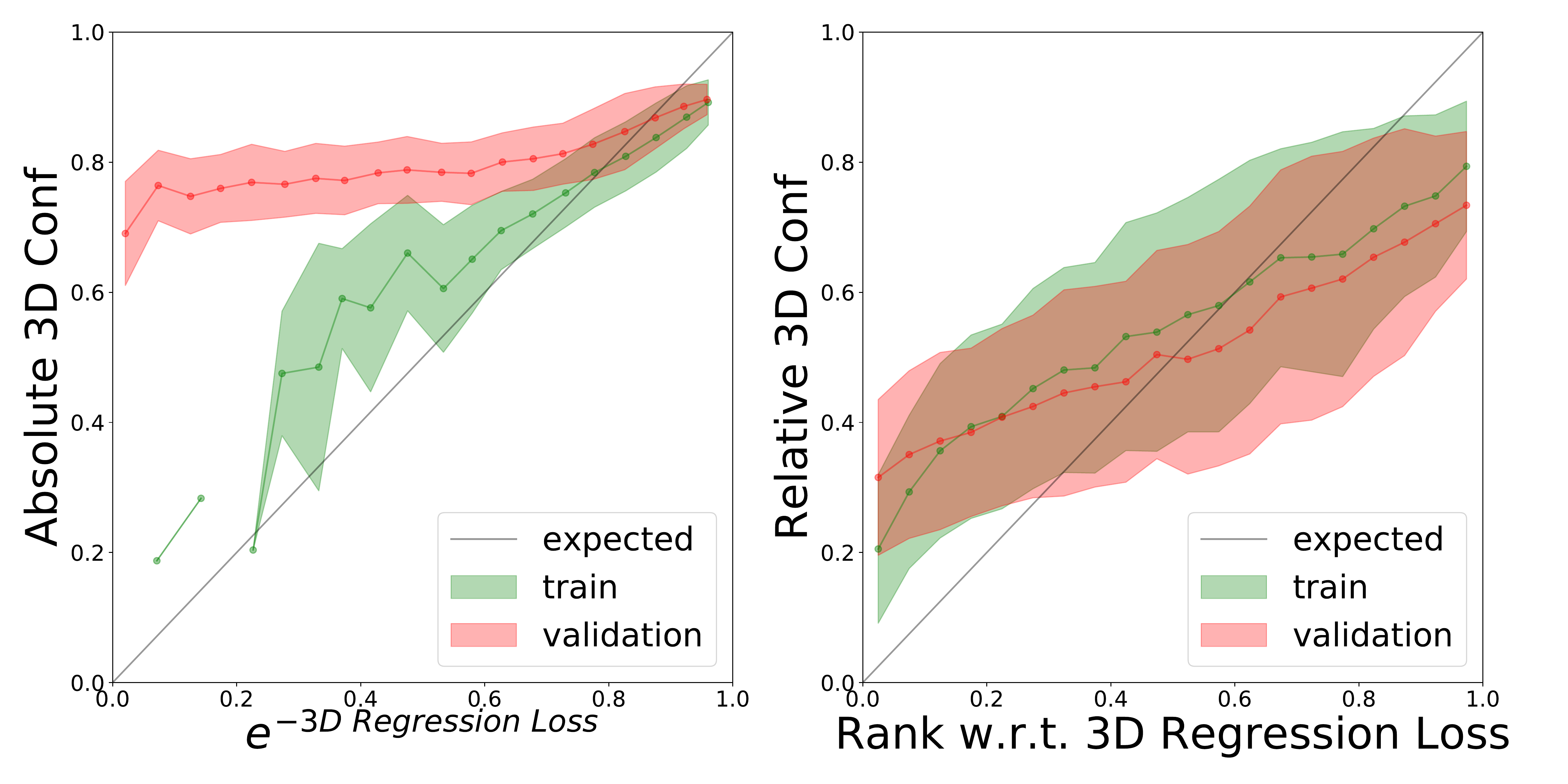}
    \caption{Binned scatter plot of the \textit{Absolute} (left) and \textit{Relative} (right) 3D confidences on the training (green) and validation (red) splits. Since the x-axis represents what the confidence is supposed to regress, the ideal curve lies on the diagonal. On the training set, both the absolute and relative confidences follow the expected curve. On the validation set however only the relative confidence aligns with the expected curve, whereas the absolute confidence consistently overestimates it, signaling the issue of being overconfident.}
\label{fig:err_bands}
\end{figure}

\myparagraph{Pros and cons of relative confidences.}
Our score takes the role of a \emph{relative} confidence, which does not give information about the quality of a prediction in absolute terms, but rather assesses the quality of the prediction relative to other predictions done by the network.
As an extreme example, a nearly perfect prediction can get zero confidence if that is the prediction that incurred the highest error in the training set. Similarly, a bad prediction can get high confidence score if it is the best prediction the network ever made. An interesting property of our confidence score is the invariance to order-preserving transformations of the losses. This renders the score more robust to scenarios where the losses change over time, which is the setting that is encountered at training time. Additionally, our confidence score does not suffer from the problem of the network becoming overconfident, because of the relative nature of our score (see Fig.~\ref{fig:err_bands}).
On the downside, an absolute confidence is typically helpful to filter predictions based on a threshold. Doing the same with a relative confidence might be cumbersome.
This is why we actually combine our relative confidence score from the 3D detection head with the 2D \emph{absolute} confidence score from the 2D detection. Indeed, the 2D detection confidence works well for the sake of removing bad quality predictions, but lacks resolution for discriminating the quality of the predictions left. This is where our relative confidence comes into play, since it does not suffer from the issue of becoming overconfident.

\begin{table}[t]
    \centering
    \resizebox{0.99\columnwidth}{!}{
    \begin{tabular}{l|ccc|ccc}    
    
        \toprule
        
        & \multicolumn{3}{c|}{\textit{Validation 3D AP}} & \multicolumn{3}{c}{\textit{Test 3D AP}} \\

        Method & Easy & Mod. & Hard & Easy & Mod. & Hard \\
         
        \midrule

        Wang et al. & 24.47 & 13.40 & 10.92 & 14.17 & 8.47 & 7.29\\ 
        + \textit{Absolute 3D Conf} & 32.44 & 20.84 & 17.26 & 18.56 & 10.99 & 9.31 \\
        + \textit{Relative 3D Conf} & \textbf{34.56} & \textbf{22.04} & \textbf{18.87} & \textbf{18.74} & \textbf{11.04} & \textbf{9.41} \\

        \midrule
        
        PatchNet &  30.53 & 17.33 & 12.80 & 15.70 & 10.15 & 8.79 \\
        + \textit{Absolute 3D Conf} &  37.04 & 23.26 & 18.78 & 22.21 & 12.51 & 10.46\\
        + \textit{Relative 3D Conf} & \textbf{38.60} & \textbf{23.68} & \textbf{19.51} & \textbf{22.40} & \textbf{12.53} & \textbf{10.64}\\

        \bottomrule
        
    \end{tabular}}
    \caption{KITTI3D Validation and Test set $AP|_{R_{40}}$ results.}
    \label{tab:kitti_ablations}
\end{table}

\section{Experiments}\label{sec:experiments}
We test the validity of our proposed 3D confidence measures by considering the deep architectures implemented in two common \pl methods, \ie the first \pl method, \textit{Wang et al.}~\cite{wang2019pseudo}, and a current top-performing state-of-the-art method, PatchNet~\cite{Ma_2020_rethink}. We modify their architectures to include our proposed \textit{3D Confidence Head}, which in all our experiments is implemented mirroring the existing \textit{3D Head}. In detail, it is implemented as one set of fully-connected layers for \cite{wang2019pseudo} and three distance-specific fully-connected modules for \cite{Ma_2020_rethink}. We follow the schedules and hyperparameters choices of \cite{Ma_2020_rethink} and ~\cite{wang2019pseudo}, with the only addition of the 3D Confidence loss which is given a weight of 1.0. Additional implementation details are given in Sec.~\ref{sec:impl_details}.
We follow the experimental protocol of all \pl-based works and evaluate our method on the KITTI3D~\cite{Geiger2012CVPR} benchmark. 
All the results presented and reported in this work have been computed with the official $AP|_{R_{40}}$ metric.

\begin{table}[tb!]
    \centering
    \resizebox{!}{0.55\columnwidth}{
    \begin{tabular}{l|ccc} 
    
        \toprule
        
        & \multicolumn{3}{c}{\textit{Car 3D AP}} \\

        Method & Easy & Mod. & Hard \\
         
        \midrule

        OFTNet~\cite{Roddick18}            	    &  1.61 &  1.32 &  1.00 \\
        FQNet~\cite{Liu+19}                     &  2.77 &  1.51 &  1.01 \\
        ROI-10D~\cite{Manhardt_2019_CVPR} 	    &  4.32 &  2.02 &  1.46 \\
        GS3D~\cite{li2019gs3d}                  &  4.47 &  2.90 &  2.47 \\
        MonoGRNet~\cite{qin2019monogrnet}  	    &  9.61 &  5.74 &  4.25 \\
        Wang et al.~\cite{wang2019pseudo}       &  9.87 &  6.40 &  5.46 \\
        MonoDIS~\cite{simo2019monodis}          & 10.37 &  7.94 &  6.40 \\
        MonoPSR~\cite{ku2019monopsr}            & 10.76 &  7.25 &  5.85 \\
        Mono3D-PL~\cite{weng2019mono3dppl}      & 10.76 &  7.50 &  6.10 \\
        SS3D~\cite{jorgensen2019ss3d}           & 10.78 &  7.68 &  6.51 \\
        MonoPair~\cite{li2020monopair}          & 13.04 &  9.99 &  8.65 \\
        SMOKE~\cite{liu2020smoke}               & 14.03 &  9.76 &  7.84 \\
        RTDM3D$^\textbf{A}$ ~\cite{li2020rtm3d} & 14.41 & 10.34 &  8.77 \\
        M3D-RPN~\cite{brazil2019m3d}       	    & 14.76 &  9.71 &  7.42 \\
        MoVi-3D~\cite{simo2020virtual}          & 15.19 & 10.90 &  9.26 \\
        PatchNet~\cite{Ma_2020_rethink}         & 15.68 & 11.12 & 10.17 \\
        AM3D~\cite{ma2019am3d}                  & 16.50 & 10.74 & 9.52 \\
        MonoDIS~\cite{simo2020pami}             & 16.50 & 12.20 & \underline{10.30} \\
        D4LCN~\cite{ding2020d4lcn}              & 16.65 & 11.72 & 9.51 \\
        Liu et al.$^\textbf{A}$ ~\cite{yuxuan2021ral} & \underline{21.65} & \textbf{13.25} & 9.91 \\
        
        \midrule
        
        Our PatchNet & \textbf{22.40} & \underline{12.53} & \textbf{10.64} \\
        
        \bottomrule
    \end{tabular}
    }
    \caption{Test set SOTA $AP|_{R_{40}}$ official results on KITTI3D for class \textit{Car}. Best scores in bold, runner-ups underlined. \textbf{A}~=~trained with additional data and only on class \textit{Car}.}
    \label{tab:kitti_test_car}
\end{table}

\begin{table}[tb!]
    \centering
    \resizebox{0.99\columnwidth}{!}{
    \begin{tabular}{l|ccc|ccc} 
    
        \toprule
        
        & \multicolumn{3}{c|}{\textit{Cyclist 3D AP}} & \multicolumn{3}{c}{\textit{Pedestrian 3D AP}} \\

        Method & Easy & Mod. & Hard & Easy & Mod. & Hard \\
         
        \midrule

        M3D-RPN~\cite{brazil2019m3d}       	    & 0.94 & 0.65 & 0.47 & 4.92 & 3.48 & 2.94 \\
        MoVi-3D~\cite{simo2020virtual}          & 1.08 & 0.63 & 0.70 & \underline{8.99} & \underline{5.44} & \underline{4.57}\\
        MonoDIS~\cite{simo2020pami}             & 1.17 & 0.54 & 0.48 & 7.79 & 5.14 & 4.42 \\
        D4LCN~\cite{ding2020d4lcn}              & 2.45 & 1.67 & 1.36 & 4.55 & 3.42 & 2.83 \\
        SS3D~\cite{jorgensen2019ss3d}           & 2.80 & 1.45 & 1.35 & 2.31 & 1.78 & 1.48 \\
        MonoPair~\cite{li2020monopair}          & \underline{3.79} & \underline{2.12} & \underline{1.83} & \textbf{10.02} & \textbf{6.68} & \textbf{5.53} \\
        
        \midrule
        
        Our - PatchNet & \textbf{7.79} & \textbf{4.32} & \textbf{3.98} & 3.00 & 1.81 & 1.59 \\
        
        \bottomrule
    \end{tabular}
    }
    \caption{Test set SOTA $AP|_{R_{40}}$ official results on KITTI3D of published multi-class methods for \textit{Cyclist} and \textit{Pedestrian}. Best scores in bold, runner-ups underlined.}
    \label{tab:kitti_test_cycped}
\end{table}

\myparagraph{Experimental results.}
In Tab.~\ref{tab:kitti_ablations} we investigate the influence of the {3D confidence} on the \pl-LiDAR methods \cite{wang2019pseudo} and~\cite{Ma_2020_rethink}. In particular, we compute the 3D object detection metrics on the validation and test set of KITTI3D with the baseline methods as well as with the addition of the \textit{3D Confidence Head} (\textit{+ 3D Confidence}) trained with both the {absolute} and {relative} learning procedure. As shown in the table, we observe a major improvement on the $3D$ $AP$. This 
validates our hypothesis about the importance of having a 3D confidence prediction component in \pl-based methods. The {relative 3D confidence} also consistently outperforms the {absolute} one, demonstrating the validity of our proposed relative formulation.
In Tab.~\ref{tab:kitti_test_car}, \ref{tab:kitti_test_cycped} we compare our results with state-of-the-art methods on the KITTI3D test set.
Our method based on PatchNet structure achieves state-of-the-art performance on the classes \textit{Car} and \textit{Cyclist}, while it does not surpasses previous approaches on the  \textit{Pedestrian}. We ascribe this behaviour to the fact that instances of the class \textit{Pedestrian} are extremely rare in the \textit{Eigen} depth training set, thus having a negative impact on the depth map quality. 

\begin{table}[t]
    \centering
    \resizebox{0.99\columnwidth}{!}{
    \begin{tabular}{l|ccc|ccc}
        \toprule
          & \multicolumn{3}{c|}{\textit{Validation} $3D~AP$} & \multicolumn{3}{c}{\textit{Test} $3D~AP$}  \\ 

        Method & Easy & Mod. & Hard & Easy & Mod. & Hard \\ 
        \midrule
        
        PatchNet                     & 20.79 & 10.55 &  8.90 & 10.88 &  7.42 & 6.51 \\
        \textit{+ Abs. 3D Conf.}     & 23.37 & 15.49 & 12.70 & 17.38 & 10.30 & 8.78\\
        \textit{+ Rel. 3D Conf.}     & \textbf{24.51} & \textbf{17.03} & \textbf{13.25} & \textbf{17.69} & \textbf{10.85} & \textbf{9.37} \\

        \bottomrule
    \end{tabular}}
    \caption{Validation and Test set $AP|_{R_{40}}$ results on KITTI3D obtained with the BTS~\cite{lee2019bts} depth estimator trained on our \textit{GeoSep} depth training split.}
    \label{tab:kitti_ablations_bias}
\end{table}

\begin{table}[t]
    \centering
    \resizebox{0.80\columnwidth}{!}{
    \begin{tabular}{c|ccc|ccc}    
    
        \toprule
        
        & \multicolumn{3}{c|}{ \textit{Wang et al. Car} 3D AP} & \multicolumn{3}{c}{\textit{PatchNet Car} 3D AP} \\

        $\beta$ & Easy & Mod. & Hard & Easy & Mod. & Hard \\
         
        \midrule

        0.1 & 28.51 & 18.78 & 15.85 & 33.69 & 21.00 & 16.42\\
        1 & 32.44 & 20.84 & 17.26 & 37.04 & 23.26 & 18.78\\
        10 & 30.72 & 19.82 & 16.43 & 32.05 & 19.06 & 15.47\\
        
        \bottomrule
        
    \end{tabular}}
    \caption{\textit{Absolute 3D Confidence} ablation results.}
    \label{tab:kitti_ablations_temp}
\end{table}

In Tab.~\ref{tab:kitti_ablations_bias} we also report the results of \textit{PatchNet + 3D Confidence} obtained by relying on a depth estimator trained on the \textit{GeoSep} depth training set. Again, this table demonstrates the merit of our contributions in term of 3D confidence estimation. Additionally, relatively to our first contribution, we notice that a gap between the validation and test set results still exists, indicating the bias issue. However, this does not invalidate our observations regarding the effectiveness of the proposed 3D confidence.

In Tab.~\ref{tab:kitti_ablations_temp} we perform a sentitivity study and analyze the behaviour of the absolute 3D confidence with respect to changes in the temperature value $\beta$. We chose a temperature of 0.1, 1 and 10 and computed results on the KITTI3D validation set. The significant change in performance demonstrates that this type of {absolute} confidence is indeed sensitive to hyperparameter tuning.

\section{Implementation details}\label{sec:impl_details}
In this section we provide details about the implementation and additional information about the hyper-parameters. Since our method is subdivided into multiple branches, we provide details of each one namely \textit{2D Detection}, \textit{Pseudo-LiDAR} and \textit{3D Detection}. In all our experiments, we trained our models on a single NVIDIA GTX 1080 Ti with 11GB of memory.

\myparagraph{2D Detection.}
As described in Sec.~\ref{sec:experiments} we do not train a 2D detector but instead rely on pre-computed 2D detections. In our experiments we used, for both validation and test set, the 2D detections used in PatchNet~\cite{Ma_2020_rethink}.

\myparagraph{Pseudo-LiDAR.}
We took the open-source code of BTS~\cite{lee2019bts} and selected the DenseNet161-based estimator. For our results on the Eigen \etal~\cite{Eigen2015} we used the model trained by the authors\footnote{\url{https://cogaplex-bts.s3.ap-northeast-2.amazonaws.com/bts_eigen_v2_pytorch_densenet161.zip}}. For the training on our novel \textit{GeoSep} splits, we used the ImageNet~\cite{Deng2009} pre-trained model and followed the official schedule and hyperparameters.

\begin{figure}[t]
    \centering
    \includegraphics[width=1.0\columnwidth]{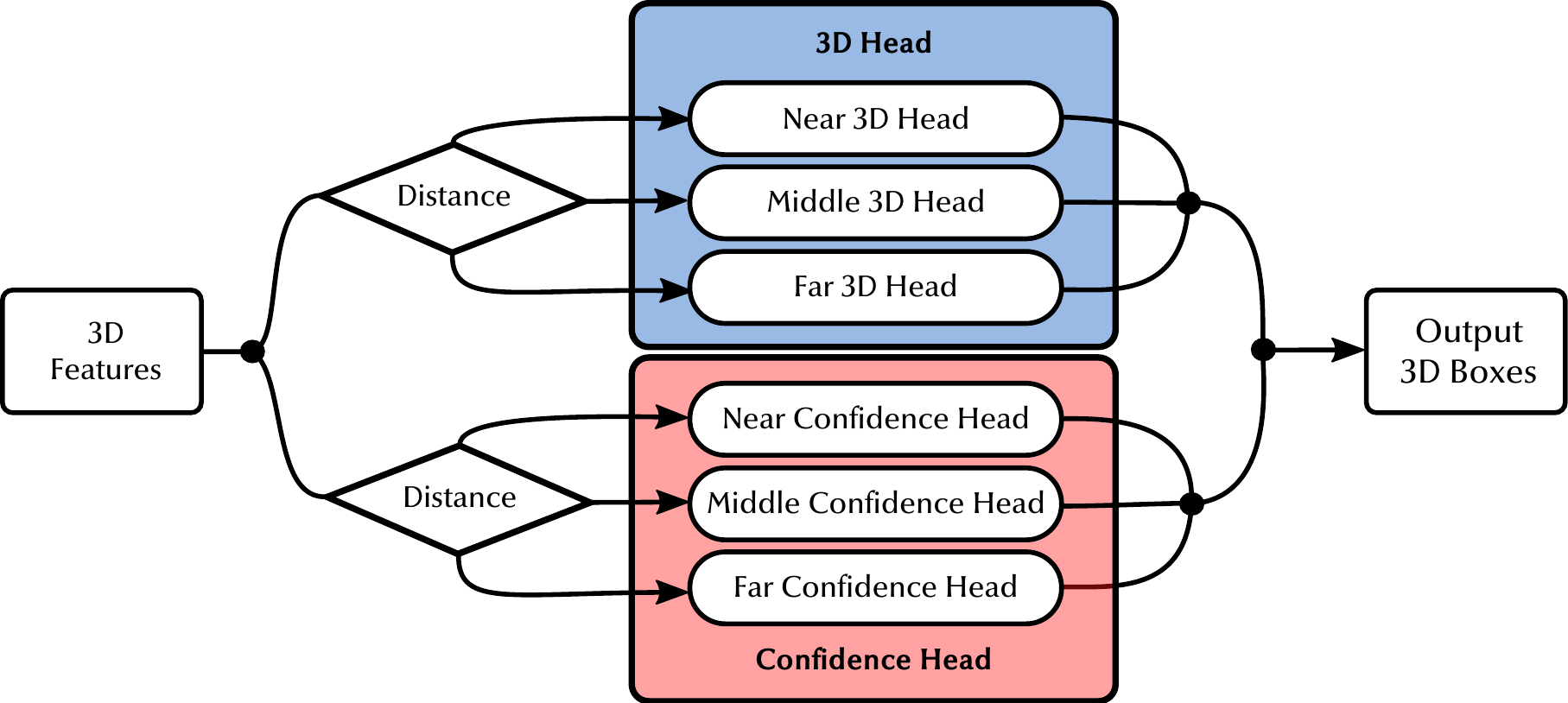}
    \caption{Example of final part of the 3d detection architecture, where we introduced our proposed \textit{3D Confidence Head} in PatchNet~\cite{Ma_2020_rethink}. The implementation of the \textit{3D Confidence Head} (red) follows the one of the \textit{3D Head} (blue).}
    \label{fig:3d_conf_head}
    \vspace{-12pt}
\end{figure}

\myparagraph{3D Detection.}
The architecture of our proposed models, \ie the ones based on Wang \textit{et al.}~\cite{wang2019pseudo} and PatchNet~\cite{Ma_2020_rethink}, always follow the official one with the only exception of the introduction of our proposed \textit{3D Confidence Head}. The implementation of this particular head closely follows the one of the respective 3D Head. In particular, for our implementation based on Wang \textit{et al.}~\cite{wang2019pseudo} we introduced a series of three fully-connected layers with \textit{512-D}, \textit{512-D}, and \textit{1-D} dimensions respectively. For the implementation of PatchNet we introduced three distance-specific heads composed by a series of three fully-connected layers with \textit{512-D}, \textit{512-D}, and \textit{1-D} dimensions respectively. We depict the PatchNet \textit{3D Head} (blue), along with our proposed \textit{3D Confidence Head} (red), in Fig.~\ref{fig:3d_conf_head}. We trained our model with the Adam optimizer with a learning rate of 0.001 and a batch size of 64 for 100 epochs, decreasing the learning rate by a factor of 0.1 at the $20^{th}$ and $40^{th}$ epoch.

\section{Additional Qualitative results}
After a initial description of the visualization method, we provide additional qualitative results of our detections on KITTI3D.

\begin{figure}[h!]
    \centering
    \resizebox{0.99\columnwidth}{!}{
        \includegraphics[width=\textwidth]{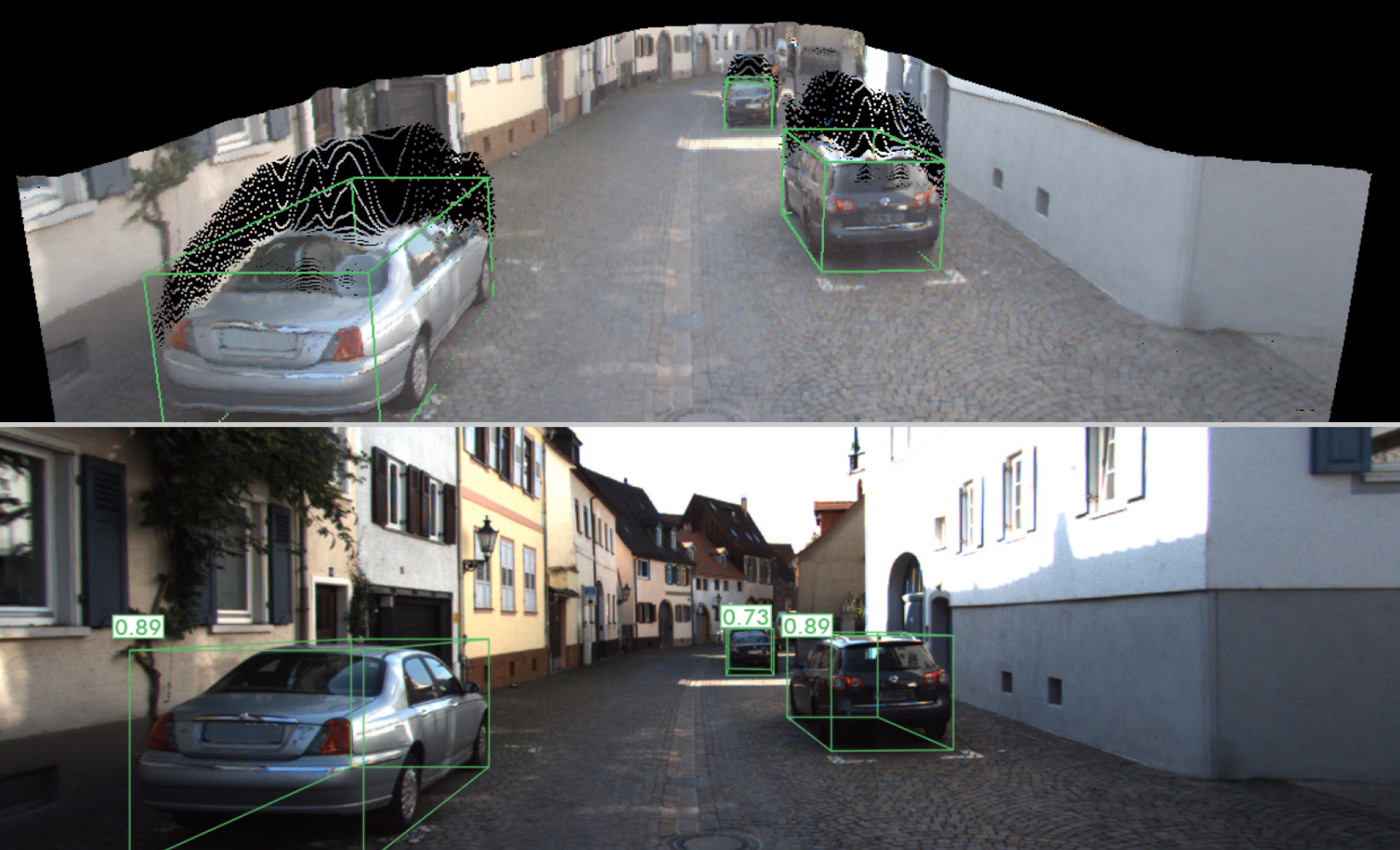}
    }
    \caption{Example of output visualization. Top: Visualization of our predictions on the colored Pseudo-LiDAR pointcloud. Bottom: Visualization of our predictions, with corresponding confidence score, on the input RGB image.}
\label{fig:video}
\end{figure}

\myparagraph{Visualization method.}
We visualize our results by super-imposing our \textit{PatchNet + Relative 3D Confidence} 3d bounding box detections on the input RGB image as well as rendered Pseudo-LiDAR pointcloud, as presented in Fig.~\ref{fig:video}.

In the top part of the images we visualize our detections on the rendered Pseudo-LiDAR pointcloud, where each point has been colored with its corresponding RGB value (if available), and consequently visualized our predicted 3d bounding-boxes in green for \textit{Car}, cyan for \textit{Cyclist} and red for \textit{Pedestrian}. The presence of black pixels (\eg on top of objects) is due to the fact that we rendered the scene from a point-of-view which is different from the one of the KITTI3D RGB camera. This change of pose inevitably introduces these black pixels on regions which were not visible from the RGB camera pose.

\begin{figure*}[h]
    \centering
    \vspace{30pt}
    \includegraphics[width=2.0\columnwidth]{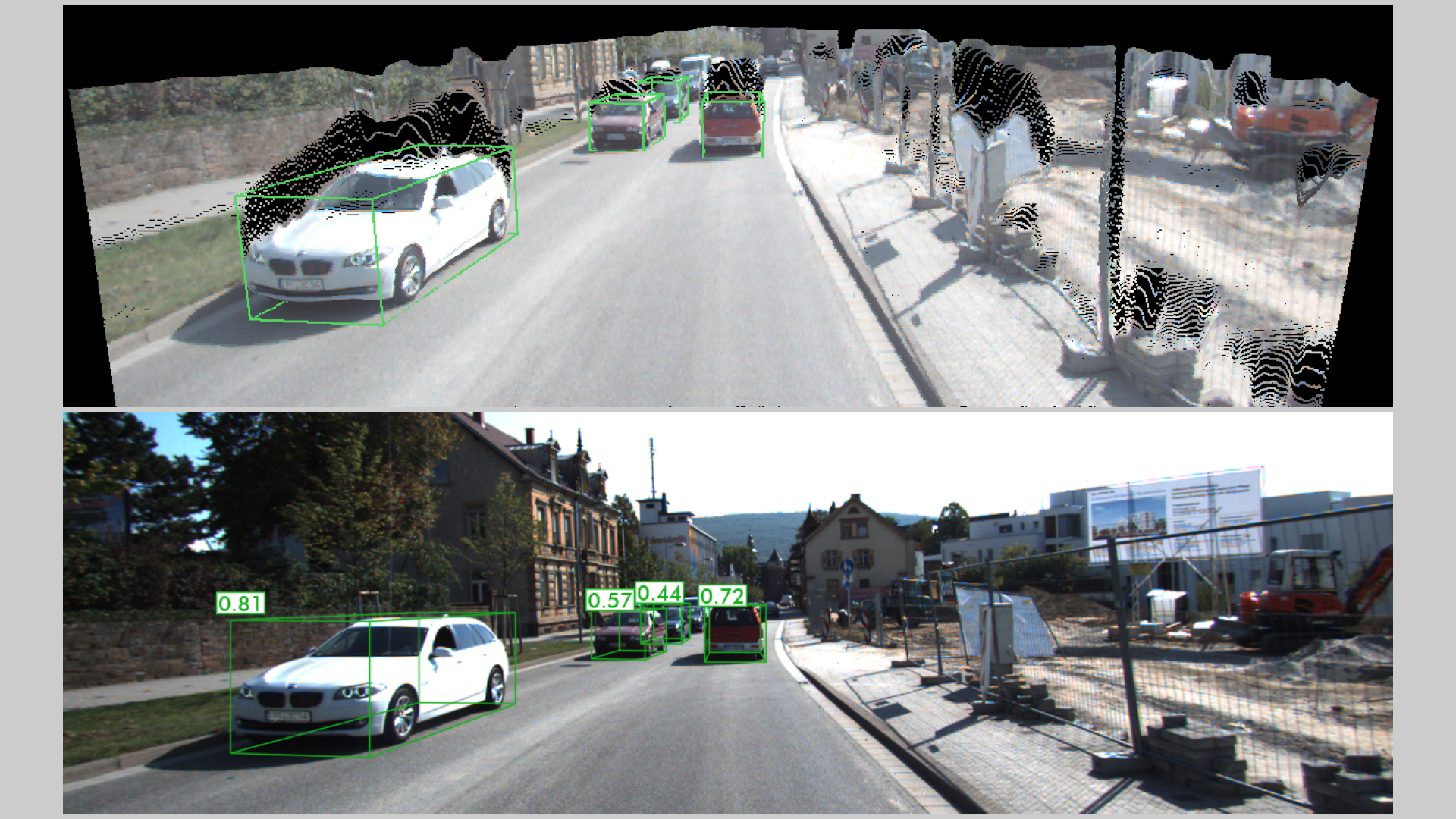}
    \includegraphics[width=2.0\columnwidth]{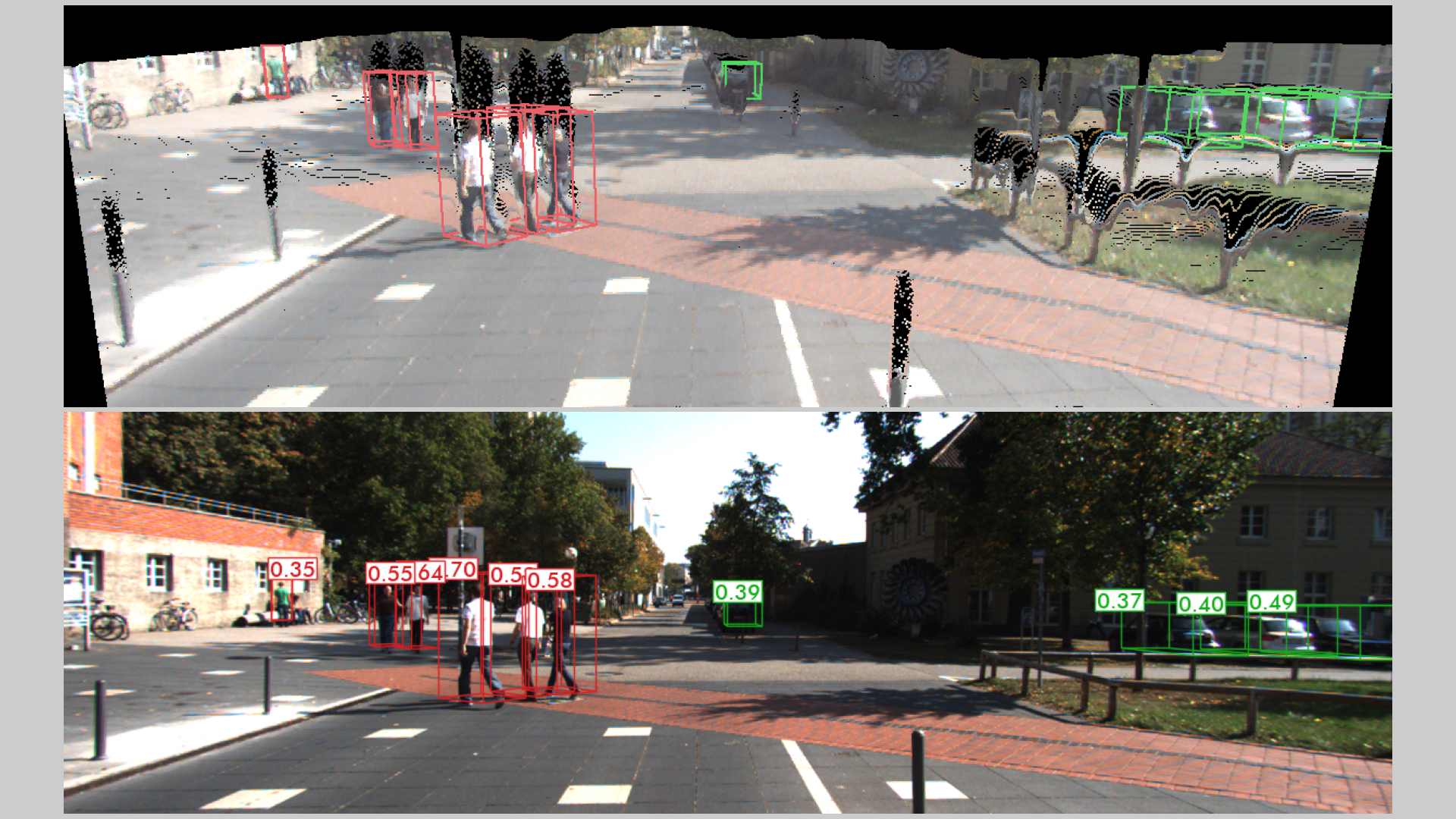}
    \caption{Additional qualitative results of our 3d bounding box detections on the KITTI3D test set.}
    \label{fig:ablation}
    \vspace{30pt}
\end{figure*}

\begin{figure*}[h]
    \centering
    \vspace{30pt}
    \includegraphics[width=2.0\columnwidth]{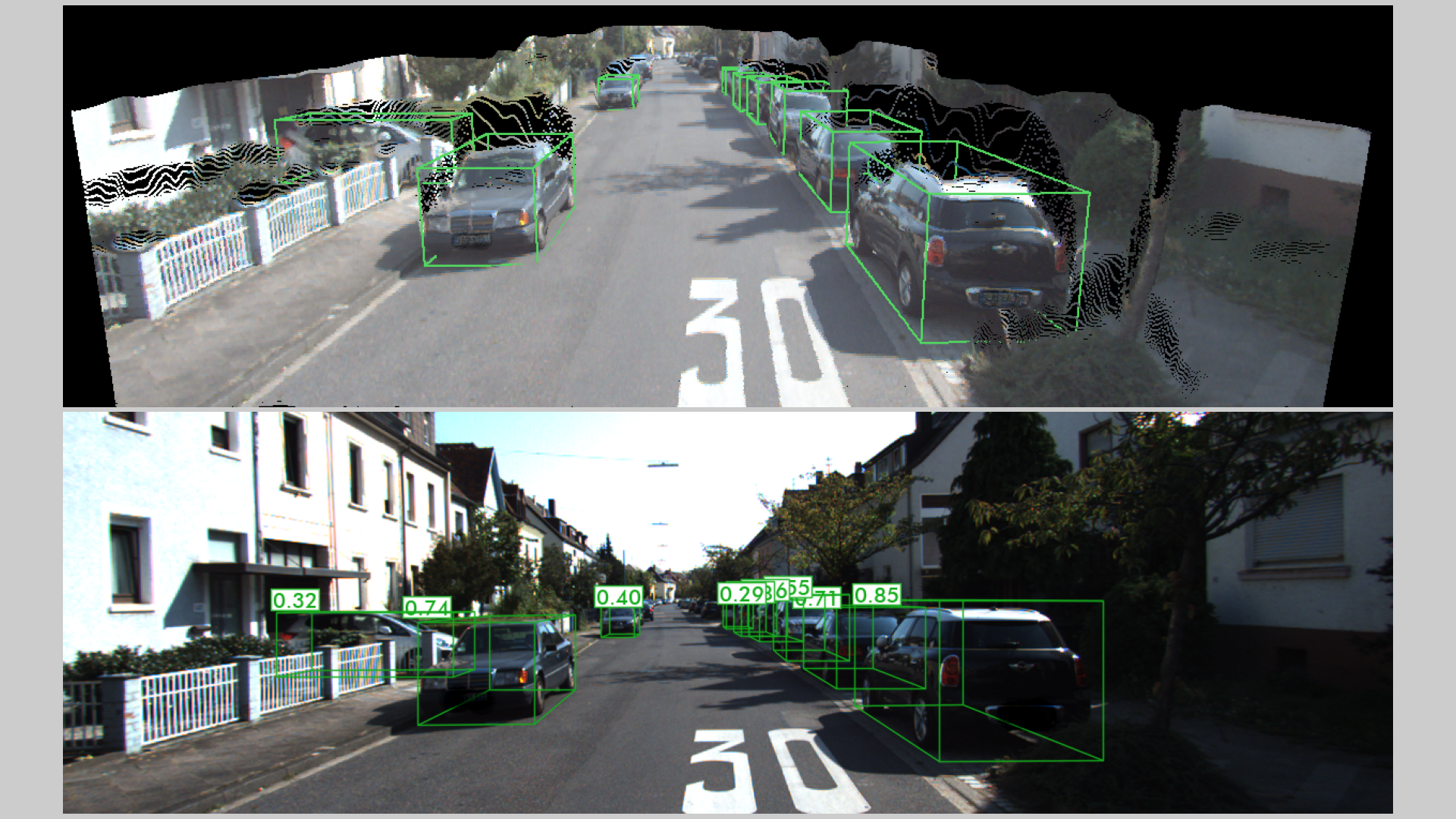}
    \\ ~ \\
    \includegraphics[width=2.0\columnwidth]{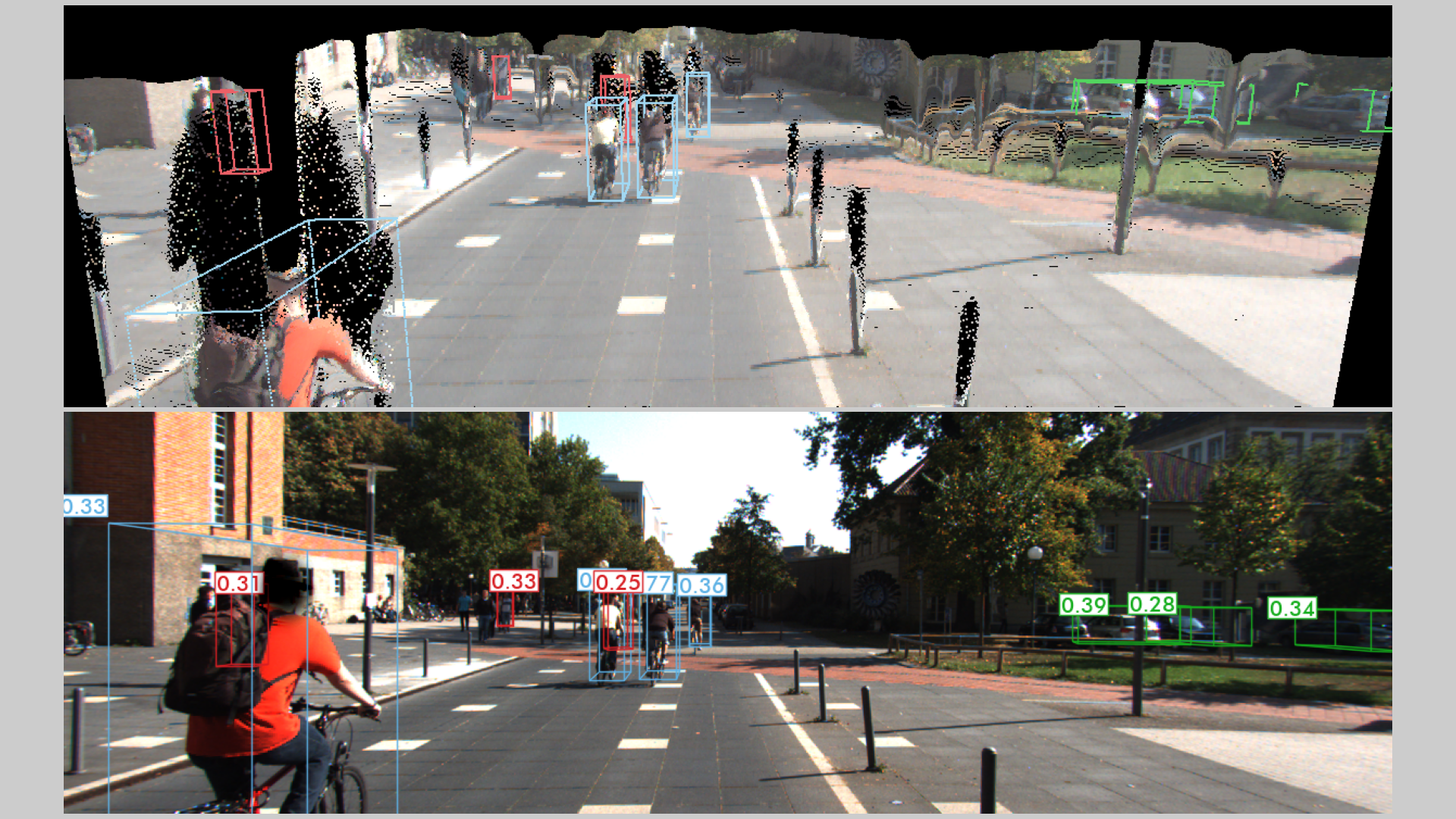}
    \caption{Additional qualitative results of our 3d bounding box detections on the KITTI3D test set.}
    \label{fig:ablation2}
    \vspace{30pt}
\end{figure*}

\myparagraph{Qualitative results on images.}
In Fig.~\ref{fig:ablation},\ref{fig:ablation2} we show our results on KITTI3D test set images. Our proposed confidence is demonstrated to reliably determine the overall quality of the predicted 3D bounding box. The confidence is in fact higher on nearer and not occluded objects, \ie where the estimation is more reliable, and seems to degrade with distance and occlusions. We also included some failure cases in which our confidence is shown to be less reliable. In particular, we have identified some imprecise or empty detections that still have fairly high confidence. 

\myparagraph{Qualitative results on video sequences.}
We further provide a qualitative video\footnote{\url{https://youtu.be/Sp8tvP6KX44}} by showing our predictions on complete KITTI3D sequences taken from the KITTI3D validation set. Unfortunately, it was not possible to provide videos on the KITTI3D test set sequences due to the unavailability of test set sequence information. The predictions are computed for each frame in an independent manner, without exploiting temporal information in any way. Despite the presence of failure cases, \eg when object are too near/far/occluded, our confidence score is shown to generally reflect the quality of the 3d detection. 


\section{Conclusions}
In this paper we have shown that top-performing Pseudo-LiDAR-based works suffer from a bias in the reported validation scores for the KITTI3D benchmark. The source of the issue is partially due to an overlap that exists between the training set used to train the upstream depth estimators, providing the depth in input to the \pl-based methods, and the validation set used for 3D object detection. 
In an attempt to validate the hypothesis we constructed an geographically separated training set for the depth estimators by ensuring geographical separation to the detection validation set. However we found that this is not sufficient to remove the bias in the validation set, which indicates the existence of a more structured nature of the issue. As a consequence, future works involving \pl-based methods on KITTI3D should avoid comparative analysis against other methods using the validation set, but rather rely on the test set.
In the second part of our work, we provided an architectural change to \pl-based methods aimed at endowing them with the ability of predicting 3D confidences. We showed that with this simple change \pl-based methods get remarkable improvements on the KITTI3D benchmark establishing a new state of the art.

\section{Acknowledgements}
We would like to thank Andreas Geiger for supporting our experiments on the KITTI3D benchmark.
We also thank Xinzhu Ma, Wanli Ouyang and Garrick Brazil for sharing their detections and for helpful discussions. 

{\small
\bibliographystyle{ieee_fullname}
\bibliography{GeneralRefs.bib}
}

\end{document}